\documentclass[runningheads]{llncs}

\usepackage[T1]{fontenc}
\usepackage{graphicx}
\usepackage{rotating}
\usepackage{array}
\usepackage{amsmath}
\usepackage{booktabs}
\usepackage[dvipsnames]{xcolor}
\usepackage{colortbl}
\usepackage{hyperref}
\usepackage[table]{xcolor}
\definecolor{lightblueA}{RGB}{220,230,250} 
\definecolor{lightblueB}{RGB}{200,220,245} 
\definecolor{BaseText}{RGB}{0,0,0}
\definecolor{InstructionText}{RGB}{0,0,204}
\definecolor{LightBox}{RGB}{230,240,255}
\usepackage{pgfpages}
\usepackage{fancyvrb}
\usepackage{fvextra}
\usepackage{helvet}
\usepackage{booktabs}
\usepackage{caption}
\usepackage{longtable}
\usepackage{listings}
\usepackage{adjustbox}
\usepackage{tcolorbox}
\usepackage{tabularx}
\usepackage{graphicx}

\usepackage{listings}
\usepackage{xcolor}
\usepackage{longtable}

\lstdefinelanguage{json}{
    basicstyle=\ttfamily\small,    
    showstringspaces=false,        
    breaklines=true,               
    frame=single,                  
    stringstyle=\color{orange},    
    keywordstyle=\color{blue},     
    morestring=[b]",               
    morekeywords={true,false,null} 
}

\definecolor{UNIFIblue}{rgb}{0.043, 0.608, 0.337}
\definecolor{backgroundUNIFI}{rgb}{1, 1, 1}

\begin{document}

\title{Hierarchical structure understanding in complex tables with VLLMs: a benchmark and experiments}

\titlerunning{Hierarchical structure understanding in complex tables with VLLMs}

 \author{Luca Bindini\inst{1}\orcidID{0009-0008-7024-2888} \and
 Simone Giovannini$^*$\inst{1}\orcidID{0009-0005-0452-8309} \and
 Simone Marinai\inst{1}\orcidID{0000-0002-6702-2277} \and
 Valeria Nardoni\inst{1}\orcidID{0009-0006-4878-3827} \and
 Kimiya Noor Ali\inst{1}\orcidID{0009-0003-5091-1603}
 }

\authorrunning{Bindini et al.}

\institute{$^{1}$DINFO -- University of Florence, Florence, Italy 
\newline
\small All authors contributed equally and are listed in alphabetical order
\newline
\small $^*$Corresponding author
\newline
\email{\{luca.bindini, simone.giovannini1, simone.marinai, valeria.nardoni, kimiya.noorali\}@unifi.it} 
}

\maketitle

\begin{abstract}
This work investigates the ability of Vision Large Language Models (VLLMs) to understand and interpret the structure of tables in scientific articles. Specifically, we explore whether VLLMs can infer the hierarchical structure of tables without additional processing.
As a basis for our experiments we use the PubTables-1M dataset, a large-scale corpus of scientific tables.
From this dataset, we extract a subset of tables that we introduce as Complex Hierarchical Tables (CHiTab): a benchmark collection of complex tables containing hierarchical headings. 
We adopt a series of prompt engineering strategies to probe the models' comprehension capabilities, experimenting with various prompt formats and writing styles. Multiple state-of-the-art open-weights VLLMs are evaluated on the benchmark first using their off-the-shelf versions and then fine-tuning some models on our task.
We also measure the performance of humans to solve the task on a small set of tables comparing with performance of the evaluated VLLMs.
The experiments support our intuition that generic VLLMs, not explicitly designed for understanding the structure of tables, can perform this task.
This study provides insights into the potential and limitations of VLLMs to process complex tables and offers guidance for future work on integrating structured data understanding into general-purpose VLLMs.

\end{abstract}

\section{Introduction}

\textit{Table Structure Recognition} (TSR) is a task of increasing interest, as it represents a fundamental first step towards the comprehensive understanding of tables within documents. Most existing TSR methods aim to reconstruct tables by either generating HTML code, identifying individual cells and their positions, inferring relationships among cells, or transforming table images into structured grid representations. However, these approaches often perform only shallow semantic analysis, frequently ignoring deep relationships among the various table elements. Recent works have introduced more advanced strategies, such as Vision Large Language Models (VLLMs) for multimodal reasoning, unified frameworks that decouple logical and physical structure recognition, and transformer-based models that jointly perform detection and parsing.

Another relevant consideration is that not all tables pose the same level of difficulty. Some exhibit very simple layouts or have clearly defined structures with fully bordered cells. According to the definition in~\cite{DBLP:journals/corr/abs-1908-04729}, \textit{if a table contains spanning cells, it is called a complicated table}, where spanning cells are defined as cells that extend across multiple rows or columns.

Building upon this definition of complexity and motivated by the need to formulate a task that takes into account the hierarchical relationships among table elements, we introduce Complex Hierarchical Tables (CHiTab), a benchmark for the recognition of hierarchical structures in complex tables.
We define a table as \textit{complex} if there exists a hierarchical relationship between the spanning cells in its header. A hierarchical relationship is established when two spanning cells are vertically aligned and the upper cell’s width fully contains the lower cell.

Starting from the annotations provided by PubTables-1M, we filter tables according to our definition and automatically generate structural annotations in a QA format, where the answer is always a single numeric value.
Once the benchmark is defined, we use CHiTab to evaluate various aspects of the behavior of state-of-the-art VLLMs when used to address our task.

The main contributions of this work can be summarized as follows:

\begin{enumerate}
    \item we propose CHiTab\footnote{\url{https://huggingface.co/datasets/AILab-UniFi/CHiTab}}, a QA-formatted TSR benchmark specifically designed to test models' deep understanding of complex table structures;
    \item using the proposed benchmark, we investigate the effect of different prompt engineering strategies on the performance of various state-of-the-art VLLMs;
    \item we evaluate these VLLMs on CHiTab in both their off-the-shelf configurations and after fine-tuning using QLoRA~\cite{DBLP:conf/nips/DettmersPHZ23}, also comparing them with a human baseline. We also investigate the stability and consistency behavior of these models.
\end{enumerate}

\section{Related work}
Understanding complex tables, those characterized by irregular layouts, merged or nested cells, and visual noise, remains a persistent challenge in document analysis. Traditional cell detection approaches often struggle in such contexts due to the absence of clear structural cues and the presence of heterogeneous visual elements~\cite{xiao2025rethinking}. These limitations are particularly pronounced in domain-specific documents, such as financial reports, where tables frequently deviate from standard formatting and are densely populated with semantically rich data~\cite{trivedi2024tabsniper}. 

To address these challenges, recent research has moved towards context-aware models that integrate semantic reasoning into the structure recognition pipeline~\cite{xiao2025rethinking}, as well as hybrid approaches that integrate traditional table parsing with semantic reasoning powered by large language models to better interpret intricate layouts~\cite{ren2025tablegpt}. In parallel, traditional deep learning approaches have continued to evolve, often benchmarking on the PubTables-1M dataset due to its scale and diversity~\cite{Smock_2022_CVPR}. 

PubTables-1M has established itself as a standard dataset for TSR, offering a rich variety of table types drawn from scientific publications. Despite its utility, the dataset also reveals fundamental issues in TSR, such as inconsistent annotations and ambiguity in defining structural elements. Standard evaluation metrics have emerged to address these issues, including Directed Adjacency Relations (DAR), Grid Table Similarity (GriTS), exact match accuracy (AccCon), and Tree Edit Distance (TED)~\cite{Smock-icdar-2023}. 

To improve TSR performance on PubTables-1M, several methods have been proposed. One approach enhances the Cascade R-CNN framework by incorporating deformable convolutions and spatial attention modules, which improve the model’s ability to handle irregular cell structures~\cite{xiao2025rethinking}. Another line of work introduces UniTabNet, a unified image-to-text model that decouples logical and physical structure recognition and uses vision–language guidance modules to improve semantic alignment~\cite{Zhang-emnlp-2024}. \ \ End-to-end pipelines based on Detection Transformers (DETR), fine-tuned on domain-specific layouts, have demonstrat\-ed robustness in handling both noisy and non-standard tables~\cite{trivedi2024tabsniper}. Additional contributions include transformer-based segmentation frameworks that simultaneously perform layout analysis, text segmentation, and structure recognition~\cite{Li-arXiv-2025}, as well as models like ClusterTabNet that apply supervised clustering strategies for joint detection and structural parsing~\cite{Polewczyk-icdar-2024}.

More recently, the introduction of VLLMs has opened up new possibilities in table understanding by enabling multimodal OCR-free reasoning on  document images. A notable benchmark in this area is MMDocBench~\cite{Zhu-corr-2024}, which evaluates the fine-grained document comprehension abilities of VLLMs across 15 tasks, including table recognition. The benchmark includes 100 samples from PubTables-1M specifically selected for assessing TSR capabilities. A framework designed to improve table recognition with VLLMs. Neighbor-Guided Tool\-chain Reasoner (NGTR)~\cite{Zhou-arxiv-2024} introduces a hierarchical evaluation setting that spans multiple levels -- cell, row, column, and full table -- offering a more granular assessment of VLLM reasoning. The authors identify input image quality as a major limiting factor in VLLM performance and respond with a pre-processing pipeline that includes lightweight enhancements such as image upscaling, binarization, and border refinement. In terms of prompting, NGTR leverages instruction-tuned prompts that guide the VLLM not only to localize table components but also to perform semantic reasoning. Additionally, the benchmark incorporates contrastive evaluation tasks to test model robustness under visually ambiguous or modified conditions.

Together, these advancements reflect a shift from purely detection-based techniques to reasoning-oriented models capable of handling the complexity of real-world tables. The integration of visual pre-processing, instruction-tuned prompting, and multimodal representation learning through VLLMs marks a promising direction for future research in TSR.

Benchmarks like MMDocBench assess cell detection or grid reconstruction, yet they do not verify whether models recover the hierarchical parent–child links created by multi‑spanning headers. As a result, there is still no public test that validates structural header understanding. CHiTab closes this gap by retaining only tables with explicit header hierarchies and casting their interpretation into concise integer‑answer QA tasks, offering a dedicated measure of a model’s ability to reason over hierarchical table layouts.

\section{CHiTab Benchmark}
\begin{figure}[t]
  \centering
  \includegraphics[width=\linewidth]{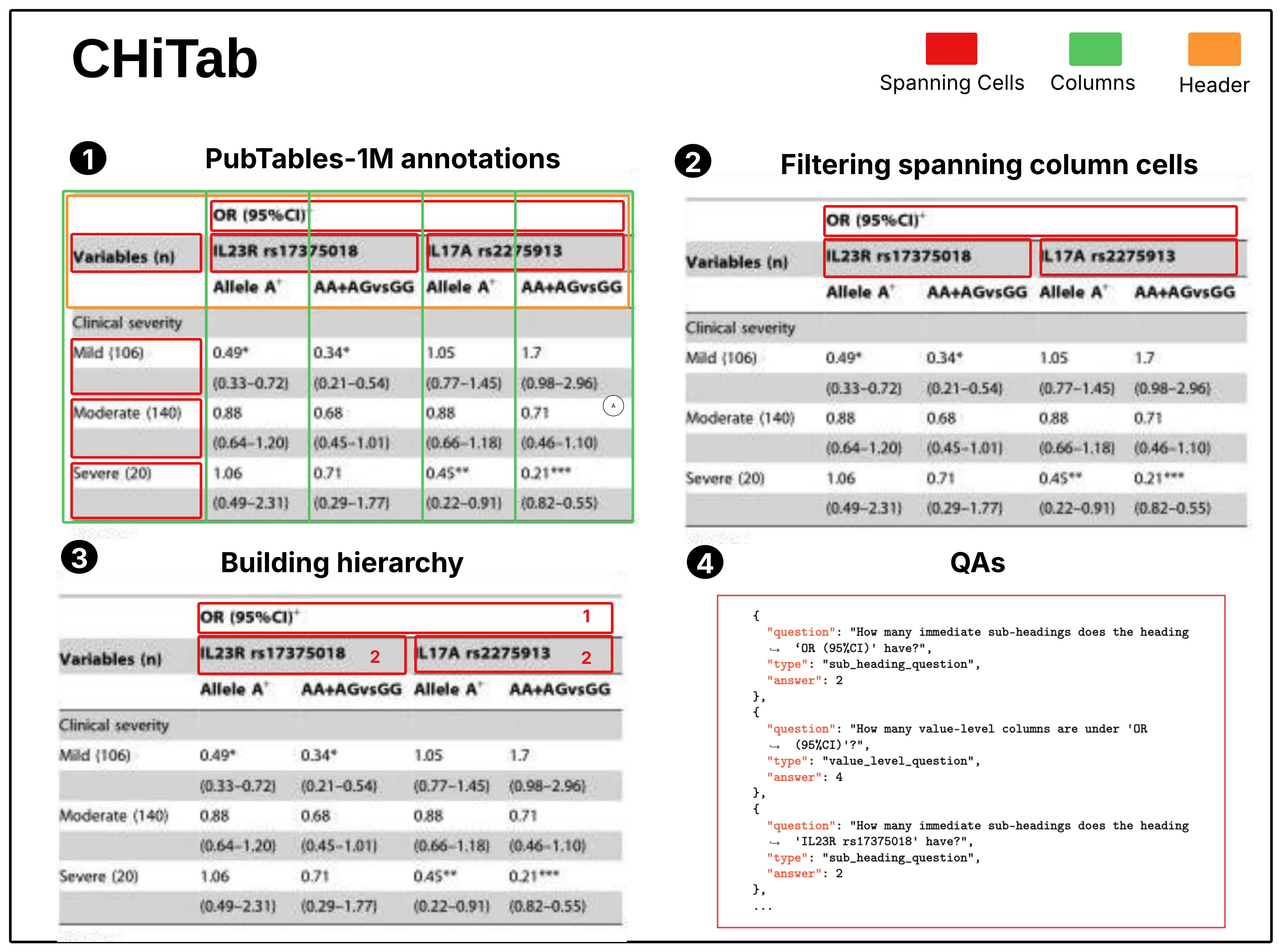}
\caption{CHiTab Benchmark outline. Steps \textcircled{\scriptsize 1} – \textcircled{\scriptsize 4} represent the pipeline stages.}
  \label{fig:outline}
\end{figure}

The CHiTab benchmark focuses on tables in the PubTables-1M corpus whose column headers form an explicit \emph{hierarchical organisation}. We select cases where one or more headings in header span multiple columns so as to express parent-child relationships between header levels. Thus, rather than including every table that merely contains a spanning cell, we retain only those in which spanning cells collectively encode a coherent header hierarchy. 
This section describes the proposed pipeline  (Fig.~\ref{fig:outline}): first, we analyze the table filtering procedure, that selects structurally challenging examples; second, we outline how header hierarchies are recovered; third, we explain the generation of ground-truth for QA tasks; finally, we present summary statistics of the resulting benchmark.

\subsection{Table Filtering}
\label{sec:table_filtering}

To construct the benchmark, we use the annotations from the PubTables-1M dataset, specifically the XML files containing structural element bounding boxes (the geometric regions corresponding to rows, columns, and cells and the accompanying JSON files with word-level annotations Fig.~\ref{fig:outline}-\textcircled{\scriptsize 1}). 
We design a script that for each table matches words to their corresponding structural components based on spatial overlap. 
A word is assigned to a structural element if its bounding box intersects with that element’s region. 
This allows us to reconstruct the full textual content of each table component by aggregating the matched words in reading order (top to bottom, left to right). 
The result is an enriched annotation format that combines layout and content information, providing a more complete representation of each table. 

To ensure that our benchmark emphasizes non-trivial structural reasoning, we apply a series of filtering steps to the full set of annotated tables; this includes all tables in the training, validation, and test splits. 
First, we select only tables that contain at least one table spanning cell. 
Then we filter out spanning cells that do not meaningfully intersect with more than one column, using a 90\% horizontal overlap threshold to eliminate marginal or noisy cases. 
Finally, we retain only those tables containing at least two spanning cells that exhibit a vertical dependency specifically, where one cell appears directly below another while sharing the same horizontal span (Fig.~\ref{fig:outline}-\textcircled{\scriptsize 2}). 

\subsection{Header Hierarchy Construction}
\label{sec:hierarchy}
In complex tables, column headers often exhibit a multi-level hierarchical structure, where high-level categories span across multiple subordinate fields. To reconstruct this structure, we apply a set of spatial heuristics over the detected spanning cells in each table, treating each table independently. 

Spanning cells are first sorted in reading order (top-to-bottom, left-to-right). For each cell, we assign a parent (if there is any) from among the candidates positioned above it in the layout. A cell is considered a valid parent if it is vertically above the candidate child, fully contains it along the horizontal axis, and is visually wider. In addition, the vertical distance between the two cells must be relatively small specifically, less than half the height of the child. When such a parent is found, we establish a directed edge in the hierarchy. The result is a forest structure, where nodes correspond to header cells and edges represent parent-child relationships in the visual and semantic hierarchy of the table 
(Fig.~\ref{fig:outline}-\textcircled{\scriptsize 3}).

\subsection{Ground Truth Generation for QA Tasks}
\label{sec:gt}
Based on the extracted header hierarchies, we generate ground truth annotations for two QA tasks:  Value-level QA ({\em VLQA}) and Sub-heading QA ({\em SHQA}) (Fig.~\ref{fig:outline}-\textcircled{\scriptsize 4}):

\paragraph{VLQA:} The goal is to count the number of value-level columns $\lvert\mathcal{I}(h)\rvert$ under a heading $h$, where $\mathcal{I}(h)$ denotes the set of value-level columns whose bounding boxes intersect $h$. This is computed counting the number of spatial intersections between the given heading and the PubTables-1M annotated columns.

\paragraph{SHQA:} The task is to count the number of direct sub-headings for a given heading. A direct sub-heading is defined as a heading that appears immediately below the given heading and is entirely contained within its spatial boundaries. The count is based on the previously built hierarchy according to Eq.~\ref{eq:gt}.

\begin{equation}
\mathrm{SHQA}(h)=
\underbrace{\lvert\mathcal{I}(h)\rvert}_{
  \substack{\text{value-level}\\\text{columns }\cap\,h}
}\;+\;
\underbrace{\lvert\mathcal{C}(h)\rvert}_{
  \substack{\text{direct}\\\text{children of }h}
}\;-\;
\sum_{c\in\mathcal{C}(h)}
\underbrace{\lvert\mathcal{I}(c)\rvert}_{
  \substack{\text{value-level}\\\text{columns }\cap\,c}
}
\label{eq:gt}
\end{equation}

\begin{table}[t!]
\begin{tcolorbox}[
    colback=LightBox,
    colframe=blue!40,
    title=\textbf{Question Phrasing Styles},
    sharp corners,
    boxrule=0.8pt,
    fonttitle=\bfseries
]
\footnotesize

\begin{adjustbox}{max width=\textwidth}
\begin{tabularx}{\textwidth}{>{\centering\arraybackslash}m{3.5cm} >{\arraybackslash}X}
    \textbf{Prompt Type} & \textbf{Example Text} \\
    \addlinespace[0.3em] \hline \addlinespace[0.3em]

    \texttt{Base Question} &
    \textcolor{BaseText}{"How many immediate sub-headings does the heading \textbf{'column\_name'} have?"} \\\\
    
    \texttt{With Explanation} &
    \textcolor{BaseText}{"How many immediate sub-headings does the heading \textbf{'column\_name'} have?"} 
    \textcolor{InstructionText}{\textit{An immediate sub-heading is one directly below the heading in reading order.}}\textcolor{BaseText}{"} \\\\

    \texttt{Uppercase} &"\textcolor{InstructionText}{\textit{HOW MANY IMMEDIATE SUB-HEADINGS DOES THE HEADING}} \textcolor{BaseText}{\textbf{'column\_name'}} \textcolor{InstructionText}{\textit{HAVE?}}\textcolor{BaseText}{"} \\\\

    \texttt{Polite} &"\textcolor{InstructionText}{\textit{Would you be so kind as to let me know}} \textcolor{BaseText}{" how many immediate sub-headings does the heading \textbf{'column\_name'} have?"} 
    \textcolor{InstructionText}{\textit{Thank you so much for your time!}}\textcolor{BaseText}{"} \\\\

    \texttt{GPT Short} &"\textcolor{InstructionText}{\textit{What is the count of direct sub-headings under the heading}} \textcolor{BaseText}{\textbf{'column\_name'}}\textcolor{InstructionText}{\textit{?}}\textcolor{BaseText}{"} \\\\

    \texttt{GPT Long} &"\textcolor{InstructionText}{\textit{Considering the hierarchical structure of the table, determine how many immediate child headings are associated with}} \textcolor{BaseText}{\textbf{'column\_name'}}\textcolor{InstructionText}{\textit{.}}\textcolor{BaseText}{"} \\\\

    \texttt{Motivation} &"\textcolor{InstructionText}{\textit{I know this is a very hard task but you can do it! Don't give up now!}} 
    \textcolor{BaseText}{" How many immediate sub-headings does the heading \textbf{'column\_name'} have?}\textcolor{BaseText}{"}\\\\

    \texttt{Reward} &"\textcolor{InstructionText}{\textit{I will give you 1000 euros if you help me with this task.}} 
    \textcolor{BaseText}{" How many immediate sub-headings does the heading \textbf{'column\_name'} have?}\textcolor{BaseText}{"} \\

\end{tabularx}
\end{adjustbox}
\end{tcolorbox}
\caption{
Prompt styles used in our study. 
\textcolor{BaseText}{Black text represents the base question,} 
\textcolor{InstructionText}{\textit{blue italic text}} \textcolor{BaseText}{indicates the prompt extension}, 
and \textbf{bold} highlights dynamic elements, in this case the column name.}
\label{tab:prompt_styles}
\end{table}

In Eq.~\eqref{eq:gt}, $\mathcal{C}(h)$ stands for the set of direct sub-headings of $h$, so $\lvert\mathcal{C}(h)\rvert$ is the number of those.  
Because each sub-headings $c\!\in\!\mathcal{C}(h)$ may itself overlap some value-level columns, the sum $\sum_{c\in\mathcal{C}(h)}\lvert\mathcal{I}(c)\rvert$ subtracts those overlaps to avoid double-counting. 
As we can infer directly from the formula, if the given heading has no children the answer of SHQA is the same as the one from VLQA.

To evaluate the robustness of VLLMs to linguistic variations, we systematically assessed multiple prompt formulations for each QA instance. These prompts differed in style and structure, encompassing variations such as: inclusion of explanatory context, usage of uppercase text, polite phrasing, prompts refined by GPT with and without explanations, encouraging language, and the addition of monetary reward (Table~\ref{tab:prompt_styles}).
This methodological approach was informed by prior research indicating that even minor alterations in prompt design can lead to significant fluctuations in model performance~\cite{leidinger2023language}.

\subsection{Benchmark Statistics}
\label{sec:benchstats}

The filtering procedure described in Section~\ref{sec:table_filtering} drastically reduces the number of tables of the original PubTables-1M that are evaluated. Table~\ref{tab:coverage_benchmark} quantifies this reduction: in every split, only about \(2.5\%\) of the initial tables satisfy all benchmark  constraints.  Nevertheless the number of filtered tables is large enough to perform interesting evaluations. 

\begin{table*}[t]
    \centering
    \footnotesize
    \renewcommand{\arraystretch}{1.5}
    \setlength{\tabcolsep}{12pt}
    \rowcolors{2}{white}{gray!15}
    \begin{tabular}{l|r|r|r}
        \textbf{Split} & \textbf{PubTables-1M} & \textbf{CHiTab} & \textbf{Coverage (\%)} \\
        \hline
        Train & 758\,849 & 18\,909 & 2.49 \\
        Valid & 94\,959  & 2\,325  & 2.45 \\
        Test  & 93\,834  & 2\,428  & 2.59 \\
    \end{tabular}
    \caption{Coverage of CHiTab benchmark w.r.t the original PubTables-1M splits.}
    \label{tab:coverage_benchmark}
\end{table*}

 Table~\ref{tab:mean_answer_both} reports the average numeric answer and its standard deviation for each question type and split. As expected, VLQA yields larger values on average because it counts all leaf columns beneath a header, whereas SHQA considers only direct children.

\begin{table*}[b]
    \centering
    \footnotesize
    \renewcommand{\arraystretch}{1.5}
    \setlength{\tabcolsep}{12pt}
    \rowcolors{2}{white}{gray!15}
    \begin{tabular}{l|r|r|r}
        \textbf{Split} & \textbf{SHQA} & \textbf{VLQA} & \textbf{\# Questions} \\
        \hline
        Train & 2.42$\pm$1.11 & 3.34$\pm$2.14 & 85\,691 \\
        Valid & 2.40$\pm$1.10 & 3.33$\pm$2.15 & 10\,519 \\
        Test  & 2.41$\pm$1.03 & 3.32$\pm$2.06 & 10\,944 \\
    \end{tabular}
    \caption{Average numeric answer value by split and question type.}
    \label{tab:mean_answer_both}
\end{table*}

\begin{figure}[t]
  \centering
  \includegraphics[width=.85\linewidth]{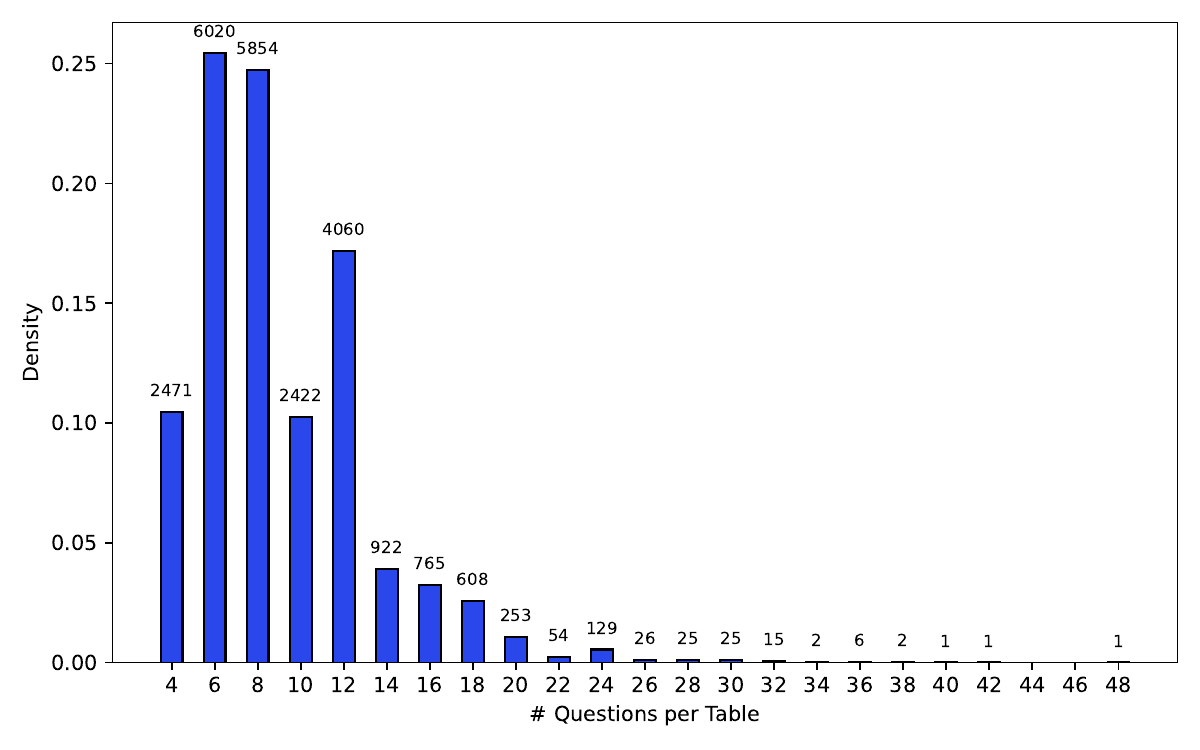}
  \caption{Histogram of the number of questions per table.}
  \label{fig:hist_qpt}
\end{figure}

Figure~\ref{fig:hist_qpt} complements these statistics by showing the distribution of questions per table (all splits combined). The peak is around 6-8 questions, yet the long tail highlights tables with deeply nested headers that generate many more questions.

\section{Experiments}

After defining CHiTab, we employ it to assess the table understanding capabilities of several popular VLLMs that currently achieve state-of-the-art performance on common DocumentAI tasks. A sample of four VLLMs is selected, chosen to represent a range of model sizes and sources, so as  to provide the broadest possible overview. The selected models are: \texttt{Granite Vision 3.2 2B}, \texttt{Qwen2.5-VL-Instruct-7B}, \texttt{Mistral Small 3.1 24B Instruct}, and  \texttt{Gemma3 27B Instruction Tuned}.  \  \ For computational reasons, \texttt{Gemma3} \ and  \ \texttt{Mistral Small 3.1} are used in their quantized \texttt{FP16} versions.

The experiments presented in the following sections aim to investigate several aspects: the effectiveness of off-the-shelf models, the inherent difficulty of the task, the impact of prompt phrasing on performance, and the consistency of model responses. The latter is analyzed by comparing the behavior of the models with that of a sample of human participants presented with the same task.

Since CHiTab consists exclusively of questions that admit a single, unambiguous numerical answer, we adopt accuracy as the sole evaluation metric. A response is considered correct only if the model returns the exact numerical value; any deviation from this is marked as incorrect. The accuracy values reported in the tables are scaled from 0 to 100.

\subsection{Prompt tuning}

Before conducting extensive experiments on the test set, we explore the optimal prompt formulation and assess its impact on the models' ability to interpret the structure of the provided tables.
To this end, we randomly sample 2,500 questions from the validation set, each associated with a distinct table: 1,250 belonging to the "immediate sub-heading" type and 1,250 to the "value-level column" type.
Once this subset is defined, we evaluate the models on it using the different prompt formulations defined in Section~\ref{sec:gt} and listed in Table~\ref{tab:prompt_styles}. The results of this analysis are reported in Table~\ref{tab:prompt-tuning}.

\begin{table*}[t]
\centering
\footnotesize
\renewcommand{\arraystretch}{1.5}
\setlength{\tabcolsep}{6pt}
\rowcolors{2}{white}{gray!15}
\resizebox{\textwidth}{!}{%
\begin{tabular}{l|cc|cc|cc|cc|cc}
\multicolumn{1}{c|}{} & \multicolumn{2}{c|}{\textbf{Granite}} & \multicolumn{2}{c|}{\textbf{Qwen}} & \multicolumn{2}{c|}{\textbf{Mistral}} & \multicolumn{2}{c|}{\textbf{Gemma}} & \multicolumn{2}{c}{\textbf{Average}} \\
\cline{2-11}
\textbf{Formulation} & \textbf{SH} & \textbf{VL} & \textbf{SH} & \textbf{VL} & \textbf{SH} & \textbf{VL} & \textbf{SH} & \textbf{VL} & \textbf{SH} & \textbf{VL} \\
\hline
\textbf{Base Question} & \textbf{55.0} & 27.3 & 52.6 & 37.5 & \textbf{54.1} & 46.3 & \underline{49.1} & 51.0 & \cellcolor{lightblueB} \textbf{52.7}  & \cellcolor{lightblueB} 40.5 \\
\textbf{W. Expl} & \underline{54.6} & 24.1 & 43.4 & 37.7 & \underline{53.0} & \underline{49.7} & 48.2 & 51.4 & \cellcolor{lightblueA} 49.8 & \cellcolor{lightblueA} \underline{40.7} \\
\textbf{Uppercase} & 51.0 & 27.3 & 51.6 & 37.6 & 47.0 & 41.2 & \textbf{49.3} & \textbf{51.7} & \cellcolor{lightblueB} 49.7 & \cellcolor{lightblueB} 39.5 \\
\textbf{Motivational} & 51.0 & 22.9 & \underline{55.2} & \textbf{40.2} & 44.7 & 39.5 & 47.2 & 48.8 & \cellcolor{lightblueA} 49.5 & \cellcolor{lightblueA} 37.9 \\
\textbf{Reward} & 53.8 & \underline{27.8} & \textbf{62.3} & \underline{39.4} & 44.2 & 36.6 & 49.0 & 49.0 & \cellcolor{lightblueB} \underline{52.3} & \cellcolor{lightblueB} 38.2 \\
\textbf{Polite} & 43.0 & 24.8 & 42.6 & 37.1 & 51.9 & \textbf{50.4} & 38.6 & \underline{51.6} & \cellcolor{lightblueA} 44.0 & \cellcolor{lightblueA} \textbf{41.0} \\
\textbf{GPT Short} & 38.7 & \textbf{29.4} & 36.7 & 37.7 & 45.0 & 25.3 & 38.2 & 48.6 & \cellcolor{lightblueB} 39.7 & \cellcolor{lightblueB} 35.3 \\
\textbf{GPT Long} & 7.6 & 27.4 & 44.9 & 35.2 & 46.9 & 17.3 & 44.8 & 50.7 & \cellcolor{lightblueA} 36.1 & \cellcolor{lightblueA} 32.7 \\
\hline
\end{tabular}
}
\caption{Prompt tuning results. Accuracy values per model and question type on a subset of the validation split.}
\label{tab:prompt-tuning}
\end{table*}

As evident from the results, SHQA questions are generally simpler than VLQA questions, as indicated by the values in the "Average" column.
Analyzing the impact of prompt formulation, it becomes clear that even subtle variations can significantly affect model performance on a given task. Consider, for instance, the SH column for Qwen: the accuracy difference between the "Reward" and "Polite" versions reaches nearly 20 percentage points, despite the minimal difference in prompt wording: the former offers a monetary reward, while the latter merely adds a polite request to the original question.

Despite testing various prompt reformulations, the high variance and inconsistent behavior across models prevent the identification of a clearly superior prompt style. Nonetheless, the original formulation yields the highest average accuracy for SH questions and ranks third for VL questions, outperforming even the "Long" version, which includes explicit definitions of the requested concepts.

Based on these findings, the original prompt formulation is selected for all subsequent testing.

\subsection{Zero shot}

Once the prompt is fixed, the four selected models are evaluated on the CHiTab test set to assess their actual effectiveness on the proposed task.
The results are summarized in Table~\ref{tab:zs+ft}.

\begin{table}[t]
\centering
\footnotesize
\renewcommand{\arraystretch}{1.5}
\setlength{\tabcolsep}{10pt}
\rowcolors{2}{white}{gray!15}
\begin{tabular}{l|c|c|c}
\textbf{Model} & \textbf{Sub-Heading QA}  & \textbf{Value-Level QA} & \textbf{Average} \\
\hline
\textbf{Granite} & 55.3 & 26.5 & \cellcolor{lightblueB} 40.9 \\
\textbf{Qwen} & 52.5 & 34.8 & \cellcolor{lightblueA} 43.7 \\
\textbf{Mistral} & 53.4 & 45.4 & \cellcolor{lightblueB} 49.4 \\
\textbf{Gemma} & 51.5 & 45.4 & \cellcolor{lightblueA} 48.5 \\
\hline
\textbf{Qwen + FT} & 78.6 & 73.0 & \cellcolor{lightblueB} 75.8 \\
\end{tabular}
\caption{Zero shot and fine-tuning results. Accuracy values per model and question type on CHiTab test split.}
\label{tab:zs+ft}
\end{table}

As previously observed in the context of Prompt Tuning, "value-level" questions are significantly more complex than "sub-heading" ones. This increased complexity arises from their higher reasoning demands and the necessity to comprehend a broader portion of the input image. Indeed, performance differences among models are substantial for value-level questions: larger models consistently outperform smaller ones, with a notable accuracy gap of approximately 19\% between Granite (2B) and Gemma (27B).
\begin{figure}[b]
    \centering
    \includegraphics[width=0.6\linewidth]{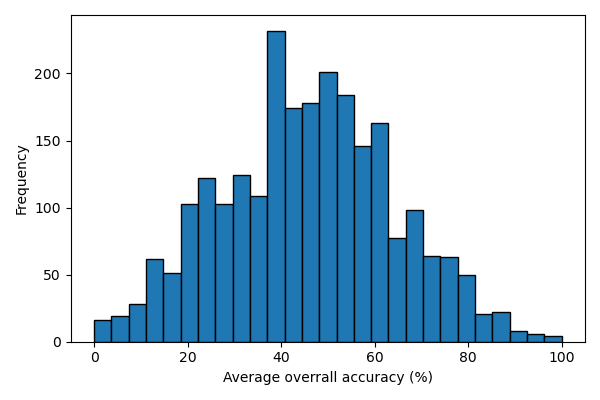}
    \caption{Distribution of the overall accuracy of the four VLLMs over the test split.}
    \label{fig:distribution}
\end{figure}
Interestingly, for sub-heading questions, the best-performing model is the much smaller Granite, achieving an accuracy of 55\%, nearly two percentage points higher than the next best model, Mistral (24B). This result likely reflects Granite's architectural design, which is optimized specifically for document understanding tasks.

\begin{figure}[t!]
    \centering
    \includegraphics[width=\linewidth]{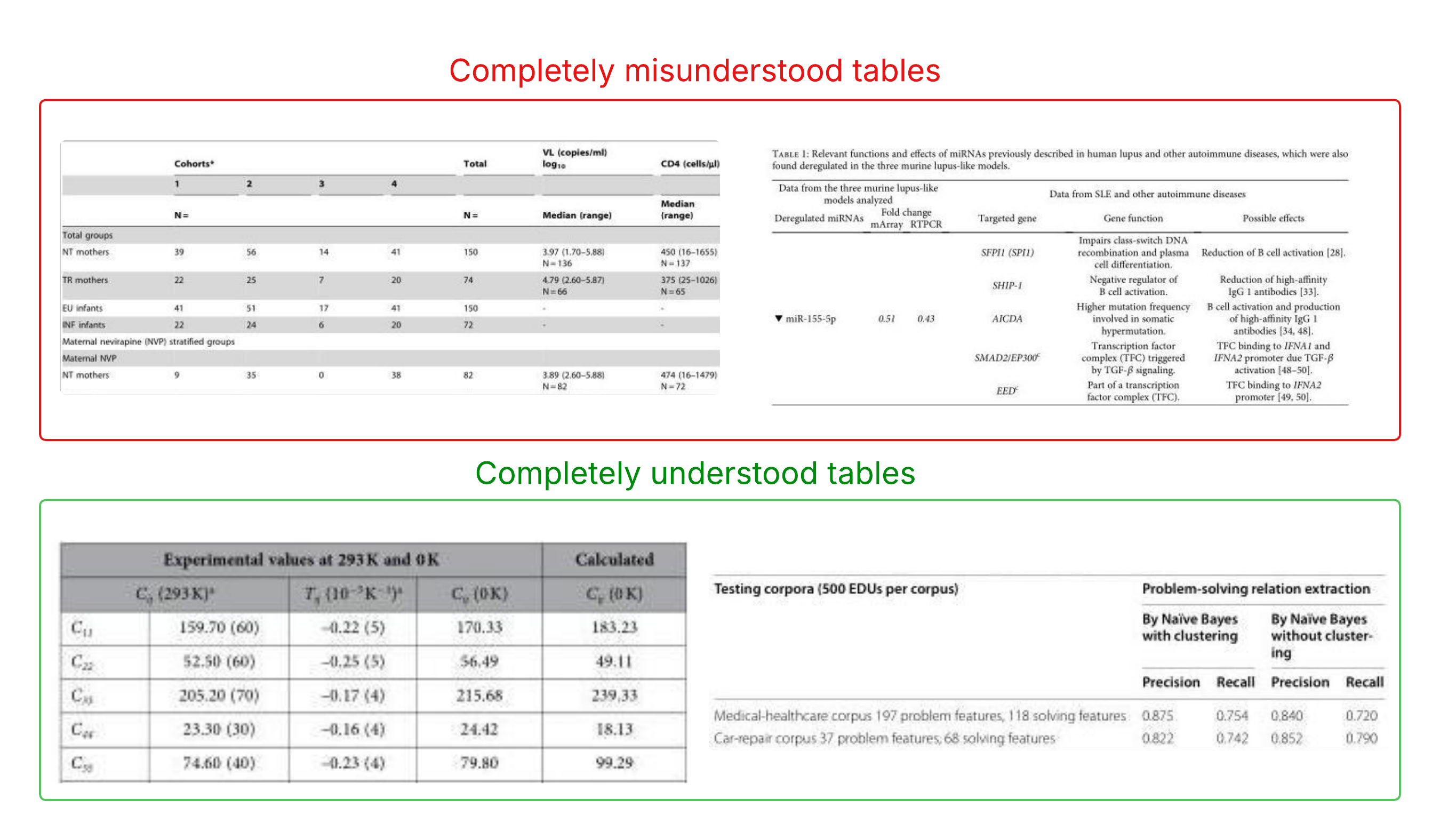}
    \caption{Examples of tables from CHiTab which are either completely misundersood or completely understood by all the tested VLLMs. All table images are shown in original resolution as found in PubTables-1M.}
    \label{fig:4examples}
\end{figure}

We further conduct an in-depth analysis of VLLMs' performance to identify which table types present the greatest challenge to the models. For each table, we compute the percentage of questions correctly answered by all four models. The distribution of these values is shown in Figure~\ref{fig:distribution}, revealing that very few tables are consistently solved across all models. Representative examples are also provided in Figure~\ref{fig:4examples}, highlighting two tables where all models failed to answer any question correctly, and two where all responses were accurate. The contrast between these table pairs is evident: the successfully resolved tables exhibit simple, well-defined structures, whereas the more challenging ones feature complex layouts with extensive spanning cells and implicit or missing border definitions.

\subsection{Fine-tuning}

Qwen2.5-VL 7B is fine-tuned using QLoRA on the provided training set. The LoRA configuration uses \texttt{r} = 8, \texttt{lora\_alpha} = 16, and \texttt{lora\_dropout} = 0.05. Training runs for a single epoch.
After fine-tuning, the model is evaluated on the test set, with results shown in Table~\ref{tab:zs+ft}. The model achieves an average accuracy improvement of over 30\% with respect to its instruct version, outperforming significantly larger models such as Mistral-Small-3.1-24B and Gemma3-27B.
These results are not intended to represent optimal performance or to suggest that the method is competitive. Rather, they provide a clearer view of the difficulty posed by the proposed benchmark.

\subsection{Human baseline}

To further assess the difficulty of the task and the complexity of the tables in CHiTab, a small subset of the dataset is presented to a sample of 29 human participants. Twenty questions were randomly sampled from the validation set, 10 of type SH and 10 of type VL, each associated with a different table. No additional explanation of the task was provided to the participants; as with the models, they were given only the table image and the question.
The same set of 20 questions is submitted to all four VLLMs evaluated in this study, with a key detail: each question is presented to each model 29 times. This experimental design enables a direct comparison between human performance and synthetic groups composed of 29 independent outputs from each model (e.g., 29 instances of Granite, 29 of Qwen, etc.). 
This setup allows us to assess not only performance but also behavioral consistency. We define consistency as a model’s ability to produce identical answers when prompted with the same input across multiple trials. A model is considered consistent for a given question if it provides the same answer in all 29 repetitions.
We acknowledge that a model’s \textit{temperature} parameter, which influences output variability, can affect behavior. To avoid introducing bias through manual configuration, we evaluate each instruct-tuned model using its default inference settings.

\begin{figure}[t]
  \centering
  \includegraphics[width=.95\linewidth]{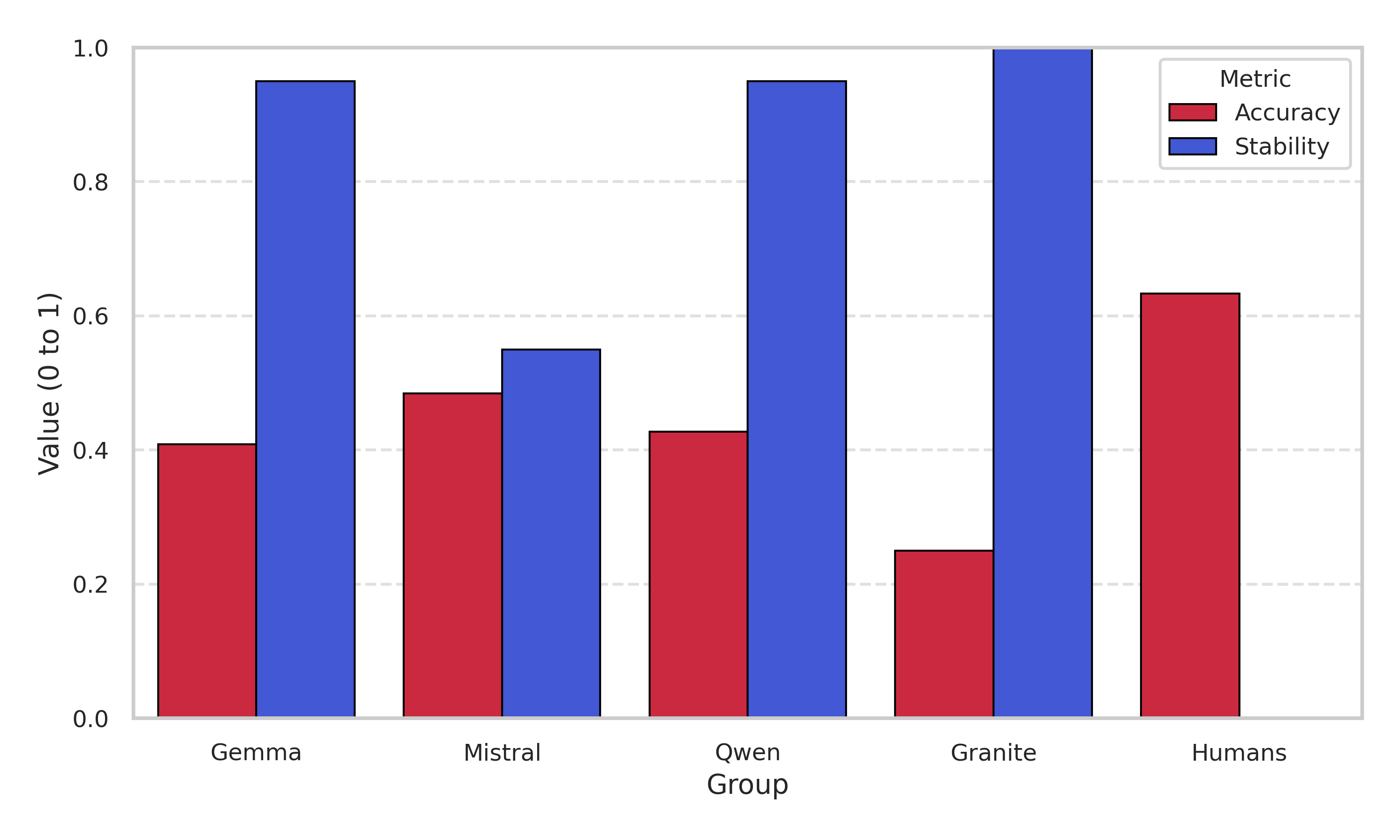}
  \caption{Accuracy and stability per group.}
  \label{fig:acc_vs_stab}
\end{figure}

The results, reported in Table~\ref{tab:model-stability} and Figure~\ref{fig:acc_vs_stab}, underscore the non-trivial nature of the task. Whether due to the intrinsic complexity of the tables or limitations in designing them, achieving deep understanding of inter-element relationships is challenging even for humans, despite their accuracy being 15 percentage points higher than the best-performing VLLM (Mistral).
An additional noteworthy finding concerns model response consistency. Despite their lower overall accuracy, Granite, Qwen, and Gemma exhibits near-perfect consistency, producing identical answers across all 29 runs for nearly every question. Only one question disrupted this unanimity for Qwen and Gemma. In contrast, Mistral, despite being the most accurate model, shows significantly less consistency, providing unanimous responses only 55\% of the time. As expected, the human group produced no unanimous answers.

\begin{table*}[t!]
\centering
\footnotesize
\renewcommand{\arraystretch}{1.5}
\setlength{\tabcolsep}{10pt}
\rowcolors{2}{white}{gray!15}
\resizebox{\textwidth}{!}{%
\begin{tabular}{l|c|c|c}
\textbf{Group} & \textbf{Mean Accuracy} & \textbf{\# Unstable Questions} & \textbf{Answer Stability (\%)} \\
\hline
\textbf{Gemma} & 41.3 & \underline{1} & \underline{95.0} \\
\textbf{Mistral} & \underline{48.1} & 9 & 55.0 \\
\textbf{Qwen} & 43.1 & \underline{1} & \underline{95.0} \\
\textbf{Granite} & 25.4 & \textbf{0} & \textbf{100.0} \\
\textbf{Humans} & \textbf{63.2} & 20 & 0.0 \\
\end{tabular}
}
\caption{Accuracy and answer stability across groups.}
\label{tab:model-stability}
\end{table*}

\section{Conclusions}

CHiTab offers a focused benchmark for evaluating how well VLLMs recover header hierarchies in complex tables.  
Zero-shot models identify about half of direct parent–child links, but struggle to count all leaf columns.  
Performance is highly prompt-sensitive, yet a light QLoRA fine-tuning on fewer than twenty thousand tables raises Qwen2.5-VL from 44\% to 76\% accuracy, showing that targeted adaptation is still crucial.

The dataset exposes current weaknesses on layouts with wide spanning cells or implicit borders, leaving a gap of roughly 15\% to human accuracy.   
We hope CHiTab spurs methods that treat tables as structured objects, closing the gap between human and machine understanding of scientific documents.

\section{Acknowledgement}
We thank the CAI4DSA actions (Collaborative Explainable neuro-symbolic AI for Decision Support Assistant) of the FAIR national project on artificial intelligence, PE1 PNRR (https://fondazione-fair.it/).
We also thank the “AI-based Service Management Suite – AISS” project.

\bibliographystyle{splncs04}
\bibliography{references}

\end{document}